\documentclass[10pt,journal,twoside]{IEEEtran}
\usepackage{amsmath,amsfonts}
\usepackage{bm}
\usepackage{algpseudocode}
\usepackage{algorithm}
\usepackage{array}
\usepackage{tabularx}
\usepackage{textcomp}
\usepackage{multirow}
\usepackage{stfloats}
\usepackage{url}
\usepackage{verbatim}
\usepackage{graphicx}
\usepackage{adjustbox}
\usepackage{mathtools}
\usepackage{makecell}
\usepackage{amssymb}
\usepackage{cite}
\usepackage{picinpar}
\usepackage{flushend}
\usepackage[utf8]{inputenc}
\usepackage{soul}
\usepackage{pifont}
\usepackage{alltt}
\usepackage[hidelinks]{hyperref}
\usepackage{enumerate}
\usepackage{siunitx}
\usepackage{epstopdf}
\usepackage{pbox}
\usepackage{dsfont}
\usepackage[table,xcdraw]{xcolor}
\usepackage{threeparttable}
\usepackage{booktabs}
\usepackage[switch]{lineno}
\usepackage{balance}
\usepackage{tikz}
\usetikzlibrary{arrows.meta,positioning,fit,backgrounds,calc}

\definecolor{arrowsteel}{RGB}{88,124,168}
\definecolor{pblue}{HTML}{4C72B0}
\definecolor{porange}{HTML}{DD8452}
\definecolor{pgreen}{HTML}{55A868}
\definecolor{pslate}{HTML}{34495E}
\definecolor{tabhead}{gray}{0.93}
\definecolor{tabbest}{RGB}{158,190,224}
\definecolor{tabsecond}{RGB}{214,228,242}
\tikzset{
  fl/.style={-{Latex[length=1.6mm,width=1.2mm]}, line width=0.6pt, draw=arrowsteel},
  bx/.style={rounded corners=1.4pt, line width=0.45pt, align=center,
             inner xsep=1.6pt, inner ysep=2.0pt, font=\sffamily\scriptsize},
  inp/.style={bx, draw=pslate!55,  fill=pslate!7},
  stg/.style={bx, draw=pblue!80,   fill=pblue!9},
  ours/.style={bx, draw=porange!90, fill=porange!18},
  res/.style={bx, draw=pgreen!80,  fill=pgreen!14},
  lbl/.style={font=\sffamily\fontsize{5.6}{6.4}\selectfont, inner sep=1pt},
}

\usepackage{newunicodechar}
\newunicodechar{−}{\ensuremath{-}}
\newunicodechar{≈}{\ensuremath{\approx}}
\newunicodechar{≤}{\ensuremath{\leq}}
\newunicodechar{≥}{\ensuremath{\geq}}
\newunicodechar{≠}{\ensuremath{\neq}}
\newunicodechar{∼}{\ensuremath{\sim}}
\newunicodechar{⇒}{\ensuremath{\Rightarrow}}
\newunicodechar{∈}{\ensuremath{\in}}
\newunicodechar{∞}{\ensuremath{\infty}}
\newunicodechar{√}{\ensuremath{\surd}}
\newunicodechar{∑}{\ensuremath{\sum}}
\newunicodechar{α}{\ensuremath{\alpha}}
\newunicodechar{β}{\ensuremath{\beta}}
\newunicodechar{γ}{\ensuremath{\gamma}}
\newunicodechar{θ}{\ensuremath{\theta}}
\newunicodechar{λ}{\ensuremath{\lambda}}
\newunicodechar{μ}{\ensuremath{\mu}}
\newunicodechar{σ}{\ensuremath{\sigma}}
\newunicodechar{φ}{\ensuremath{\phi}}
\newunicodechar{ω}{\ensuremath{\omega}}

\hypersetup{
    pdfauthor={},
    pdfsubject={},
    pdfkeywords={},
    colorlinks=true,
    linkcolor=blue,
    filecolor=magenta,
    urlcolor=blue,
}

\def\BibTeX{{\rm B\kern-.05em{\sc i\kern-.025em b}\kern-.08em
    T\kern-.1667em\lower.7ex\hbox{E}\kern-.125emX}}

\newif\ifrtmark
\rtmarkfalse
\newcommand{\rt}[1]{\ifrtmark\textcolor{red}{#1}\else#1\fi}

\DeclareMathOperator*{\argmin}{arg\,min}

\newcommand{\fL}{\mathrm{L}}            
\newcommand{\fC}{\mathrm{C}}            

\newcommand{\fB}{\mathrm{B}}            

\newif\ifanonymous
\anonymousfalse     

\ifanonymous
  \providecommand{\codeurl}{}
  \providecommand{\camerauthorblock}{}
  \providecommand{\cameraacknowledgment}{}
\else
  \IfFileExists{camera_identity.tex}{
\newcommand{\codeurl}{https://github.com/JokerJohn/P2Calib.git}

\newcommand{\camerauthorblock}{%
  \author{Xiangcheng~Hu,~\IEEEmembership{Student~Member,~IEEE}%
  \thanks{X. Hu is with the Department of Electronic and Computer Engineering,
  The Hong Kong University of Science and Technology, Clear Water Bay,
  Kowloon, Hong Kong SAR, China (e-mail:
  \texttt{xhubd@connect.ust.hk}).}%
  }
  \markboth{Submitted to IEEE Transactions on Automation Science and Engineering}%
  {HU: P$^2$CALIB: UTILIZING PATTERN PRIORS FOR LIDAR-CAMERA EXTRINSIC CALIBRATION}
}

\newcommand{\cameraacknowledgment}{%
\section*{Acknowledgment}
The authors thank Shenzhen Fortsense Technology Co., Ltd.\ for providing the
experimental equipment and data, and Shiyang Chen of X Square Robot for
productive discussions. AI-assisted tools were used to polish the writing.
}
}{%
    \GenericWarning{}{camera_identity.tex missing -- author block omitted}%
    \providecommand{\codeurl}{}
    \providecommand{\camerauthorblock}{}
    \providecommand{\cameraacknowledgment}{}}
\fi

\newcommand{\codelink}{\ifanonymous\else{} at \url{\codeurl}\fi}

\newcommand{\inputgenerated}[1]{%
  \IfFileExists{#1}{\input{#1}}{%
    \begin{table}[t]\centering\footnotesize
    \fbox{\parbox{0.9\linewidth}{\centering\textbf{Missing generated table}\\
    \texttt{\detokenize{#1}}\\[2pt]
    Run the corresponding \texttt{scripts/report\_*.py} to regenerate it.}}
    \end{table}%
    \GenericWarning{}{Generated table #1 not found -- placeholder inserted}}}

\makeatletter
\newenvironment{NoteToPractitioners}{%
  \par\addvspace{0.5\baselineskip}%
  \normalfont\@IEEEabskeysecsize\bfseries\textit{Note to Practitioners}---\relax
  \normalfont\@IEEEabskeysecsize\bfseries}%
  {\par\vspace{0.67ex}\normalfont\normalsize}
\makeatother

\begin{document}
\bstctlcite{IEEEexample:BSTcontrol}

\title{P$^2$Calib: Utilizing Pattern Priors for\\ LiDAR-Camera Extrinsic Calibration}

\ifanonymous
  \author{}
  \markboth{Submitted to IEEE Transactions on Automation Science and Engineering}%
  {Submitted to IEEE Transactions on Automation Science and Engineering}
\else
  \camerauthorblock
\fi

\maketitle

\begin{abstract}
Target-based LiDAR-camera extrinsic calibration is a prerequisite
for multi-sensor fusion in robotics. However, in the widely adopted
four-hole pipeline, \rt{calibration accuracy is limited by hole-center
extraction in the LiDAR side, where sparse angular coverage and
mixed-pixel returns displace the estimated centers}. This paper presents
P$^2$Calib, which exploits \textit{pattern priors}, geometric
constraints specified by the CAD model of the target board, to
improve calibration accuracy. First, we incorporate the known hole
radius as a fitting constraint to prevent center estimates from
degrading under sparse angular coverage. Building on the improved
hole estimates, we further enforce the rigid rectangular layout of
the four holes as a global consistency constraint to correct
residual errors across holes. Both priors are integrated into an
\rt{interactive tool that runs the pipeline from target detection to the
final extrinsic}. Experiments on simulated and real datasets
show that P$^2$Calib reduces the joint registration residual by
90\% and 82\% and the held-out reprojection error by 96\% and 77\%
over the baseline.
\rt{We will release the code and data\codelink{} to facilitate future
research.}
\end{abstract}

\begin{NoteToPractitioners}
This paper was motivated by the problem of calibrating a low-cost
solid-state LiDAR \rt{to} a camera, but it also applies to other LiDARs
that sample a target sparsely, including mechanical scanning LiDARs
with
few scan lines. Four-hole boards are widely adopted for this task because
\rt{one view of the board is enough to recover the extrinsic parameters}.
Existing approaches fit each hole independently from the point cloud
alone, so accuracy degrades when few points fall on a hole boundary, and
the \rt{practitioner} is left to acquire further views or to move the board closer
without knowing whether the result has improved. This paper suggests
exploiting the pattern of the board, whose dimensions are already
specified by its \rt{CAD model}. A radius prior fixes the diameter
of each hole during extraction, and a layout prior fixes the rectangle
formed by the four hole centers. Both priors are implemented in an
interactive tool that guides the \rt{practitioner} through the full pipeline, from
target detection to the final extrinsic, and experiments on simulated and
real data show that \rt{the recovered hole centers and the extrinsic computed
from them are more accurate and more repeatable, and that a calibration is
obtained in configurations where existing pipelines fail to detect the
board}. The
approach assumes a board manufactured to its nominal dimensions with a
rectangular four-hole layout, and a warped or out-of-tolerance board will
not benefit. In future research we will extend the pattern priors to other
target layouts and to boards whose dimensions are measured after
manufacture.
\end{NoteToPractitioners}

\begin{IEEEkeywords}
LiDAR-camera extrinsic calibration, target-based calibration,
geometric prior, constrained estimation, solid-state LiDAR.
\end{IEEEkeywords}

\IEEEpeerreviewmaketitle

\section{Introduction}
\label{sec:intro}

\subsection{Motivation and Challenges}

\IEEEPARstart{E}{xtrinsic} calibration between LiDAR and camera
estimates the rigid transform that aligns their coordinate frames
and is a prerequisite for multi-sensor fusion in robotics and
\rt{automation, including autonomous navigation, dense mapping,
industrial inspection, and automated
assembly}~\cite{xu2022fast,jiao2021robust,lv2023continuoustime,hu2024paloc,wu2021globallyoptimal,wu2022simultaneous}.
In laboratory and production settings, target-based calibration
remains the most widely adopted approach because correspondences
are known a priori and outlier rates are
low~\cite{zheng2025fastcalib,beltran2022automatic}, whereas
targetless methods inherit their accuracy from whatever structure
the environment happens to
provide~\cite{pandey2015automatic,yuan2021pixel}.
Among existing target designs, the four-hole board with ArUco
markers~\cite{zheng2025fastcalib} has gained popularity for its
simplicity: a single co-registered LiDAR scan and image suffice
to establish four 3D--3D correspondences for closed-form extrinsic
estimation.
In this pipeline, hole centers are extracted from the image via
ArUco detection and PnP with high accuracy, and the subsequent
singular value decomposition (SVD) registration is a closed-form solver that introduces no additional
error.
Calibration accuracy is therefore governed by how precisely hole
centers are extracted from the LiDAR point cloud on the board
surface.
When sufficient point density is available, standard circle fitting
yields estimates whose error is dominated by range noise.
\begin{figure}[!t]
\centering
\includegraphics[width=0.95\linewidth]{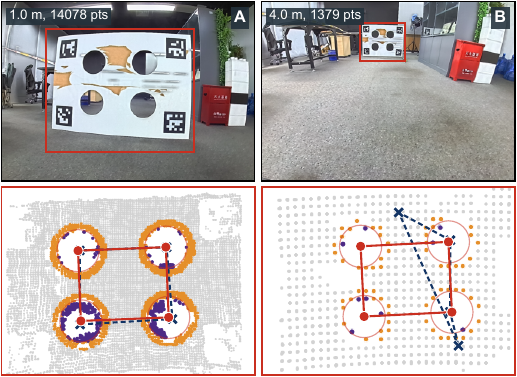}
\caption{\rt{Hole extraction with and without the pattern priors on two
scenes of one board, shown in the board plane. (\textbf{A}) A near scene
at 1.0\,m and (\textbf{B}) a far scene at 4.0\,m; each column shows the
camera image (\textbf{top}) and the board-plane points of the same scene
(\textbf{bottom}). Mixed pixels (violet) and sparse boundary sampling
(orange) displace the baseline centers (dashed) from the CAD rectangle,
which P$^2$Calib (solid) recovers.}}
\label{fig:teaser}
\vspace{-0.1cm}
\end{figure}
However, across a wide range of practical configurations, the
number of points that fall on a single hole boundary is small.
A mechanical LiDAR with few scan lines, or any LiDAR placed at a
long standoff distance, resolves the board with an angular sampling
too coarse to trace hole boundaries densely, and a small board
\rt{places} the hole aperture near that sampling limit.
\rt{Whatever the sensor, each hole is then observed as a short,
unevenly covered arc}
(Fig.~\ref{fig:teaser}).
Under such conditions, LiDAR hole-center extraction becomes the
accuracy bottleneck of the entire calibration pipeline.
We identify two challenges underlying this bottleneck:
\begin{enumerate}
    \item \textbf{Hole fitting degeneracy.}
    Under sparse angular coverage, fitting a free circle to each
    hole is ill-conditioned: a radial shift of the center can be
    compensated by an equal change in
    radius~\cite{alsharadqah2009error}, and
    mixed-pixel
    returns~\cite{tuley2005artifacts} bias
    the boundary inward along this same degenerate direction,
    amplifying center error well beyond the raw noise level
    (violet returns in Fig.~\ref{fig:teaser}).

    \item \textbf{Lack of inter-hole coupling.}
    The four circles are estimated independently with twelve free
    parameters, yet their centers must form a rigid placement of the
    machined
    rectangle~\cite{guindel2017automatic,beltran2022automatic,domhof2019extrinsic,yan2023joint,zheng2025fastcalib,jiang2025circular}.
    Where this relationship is \rt{exploited} at all, it enters after the
    fit, as a combinatorial
    check~\cite{guindel2017automatic,beltran2022automatic} or a
    rigid template
    overwrite~\cite{domhof2019extrinsic,domhof2021joint}, never as
    a constraint within the estimation itself
    (dashed quadrilateral in Fig.~\ref{fig:teaser}).
\end{enumerate}

The calibration board is manufactured with known dimensions: the
hole radius and the rectangular arrangement of the four centers are
specified exactly by the CAD model of the board.
These geometric constraints, which we collectively term
\textbf{pattern priors}, address the two challenges above yet remain
unexploited by current calibration pipelines.
For the four-hole board, they take the form of a radius prior and a
layout prior.

\subsection{Contributions}

We claim the following contributions:
\begin{itemize}
    \item We incorporate a \textbf{radius prior}
    (Section~\ref{sec:radius_prior}) that fixes the fitting radius
    to the CAD value, eliminating the center--radius
    degeneracy under sparse angular coverage.

    \item Building on the hole estimates, we enforce a
    \textbf{layout prior} (Section~\ref{sec:rigid_rect}) that
    replaces the independence assumption across holes with a global
    rigidity constraint, reducing the degrees of freedom from
    twelve to four.

    \item We integrate both priors into an \textbf{interactive
    calibration tool} (Section~\ref{sec:impl}) and evaluate the full
    method on simulated and real data \rt{(Section~\ref{sec:exp})}.
\end{itemize}

\subsection{Organization}

Section~\ref{sec:rw} reviews related calibration methods and
Section~\ref{sec:pre} formulates the four-hole pipeline.
Section~\ref{sec:method} presents the two pattern priors and
Section~\ref{sec:theory} analyzes their effect on the estimation problem.
Section~\ref{sec:impl} describes the calibration tool,
Section~\ref{sec:exp} reports the experiments, and
Section~\ref{sec:conclusion} concludes the paper.

\begin{figure*}[!h]
\centering
\begin{tikzpicture}[
  bx/.append style={font=\sffamily\scriptsize, inner xsep=2pt, inner ysep=2pt,
                    text width=26mm},
  lbl/.append style={font=\sffamily\scriptsize}]

\node[inp] (img) at (0.00, 1.15)
      {\includegraphics[height=16mm]{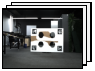}\\[1pt]Images};
\node[stg] (tag) at (3.30, 1.15)
      {\includegraphics[height=16mm]{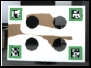}\\[1pt]Tag extraction};
\node[stg] (pose) at (6.60, 1.15)
      {\includegraphics[height=16mm]{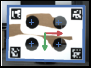}\\[1pt]Board pose \eqref{eq:campose}};
\node[stg] (pnp) at (9.90, 1.15)
      {\includegraphics[height=16mm]{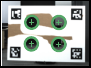}\\[1pt]Hole centers \eqref{eq:campose}};

\node[inp] (cloud) at (0.00,-1.15)
      {\includegraphics[height=16mm]{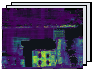}\\[1pt]Point clouds};
\node[stg] (plane) at (3.30,-1.15)
      {\includegraphics[height=16mm]{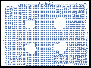}\\[1pt]Plane fitting \eqref{eq:backproject}};
\node[ours,text width=26.5mm] (radius) at (6.60,-1.15)
      {\includegraphics[height=16mm]{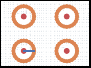}\\[1pt]\mbox{\fontsize{7}{8}\selectfont Radius prior (Sec.~\ref{sec:radius_prior})}};
\node[ours] (layout) at (9.90,-1.15)
      {\includegraphics[height=16mm]{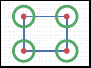}\\[1pt]Layout prior (Sec.~\ref{sec:rigid_rect})};

\node[res, minimum height=36mm, text width=28mm] (reg) at (13.55,-0.20)
      {\includegraphics[height=19mm]{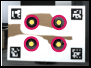}\\[2pt]Calibration \eqref{eq:extrinsic}\\[1pt]$\bm{R}^{\star}_{\fC\fL},\,\bm{t}^{\star}_{\fC\fL}$};

\draw[fl] (img)  -- (tag);
\draw[fl] (tag)  -- (pose);
\draw[fl] (pose) -- (pnp);
\draw[fl] (cloud) -- (plane);
\draw[fl] (plane) -- (radius);
\draw[fl] (radius) -- (layout);

\draw[fl] (pnp.east)    -- (reg.west |- pnp.east);
\draw[fl] (layout.east) -- (reg.west |- layout.east);

\node[lbl, above] at ($(pnp.east)!0.5!(reg.west |- pnp.east)+(0,0.04)$) {$\bm{p}^{\fC}_k$};
\node[lbl, above] at ($(layout.east)!0.5!(reg.west |- layout.east)+(0,0.04)$) {$\bm{p}^{\fL}_k$};

\begin{scope}[on background layer]
  \node[draw=pslate!35, fill=pslate!3, rounded corners=3pt, line width=0.5pt,
        inner sep=3.2mm, fit=(img)(cloud)(reg)] (frame) {};
\end{scope}
\node[anchor=north east, font=\sffamily\large\bfseries, text=pslate!85,
      inner sep=1.6mm] at (frame.north east) {P$^2$Calib};
\end{tikzpicture}
\caption{P$^2$Calib pipeline with the data at each
stage. The camera branch (\textbf{top}) detects the ArUco tags, recovers
the board pose and obtains $\bm{p}^{\fC}_k$; the LiDAR branch
(\textbf{bottom}) extracts the
board plane and estimates $\bm{p}^{\fL}_k$ under the two priors
\rt{marked in orange; closed-form registration returns the extrinsic}.}
\label{fig:overview}
\vspace{-0.1cm}
\end{figure*}
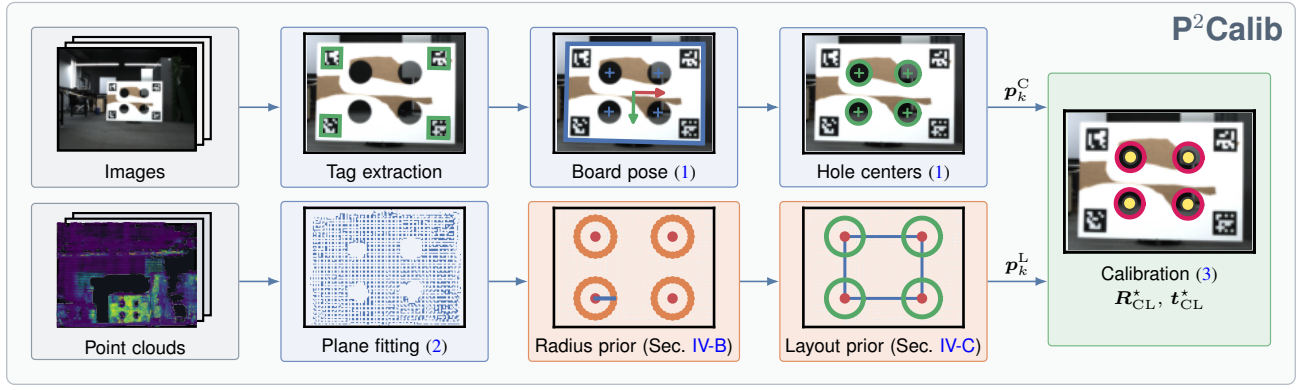

\section{Related Work}
\label{sec:rw}

LiDAR--camera extrinsic calibration methods are commonly grouped
into targetless, motion-based, and target-based
categories~\cite{zheng2025fastcalib,taylor2016motion,persic2021spatiotemporal}.
We briefly review the first two to motivate the target-based
setting, then examine how target-based methods \rt{exploit} the geometry of
the calibration board, which is where this work is positioned.

\subsection{Targetless and Motion-Based Methods}

Targetless methods replace a physical target with signals harvested
from the scene: statistical approaches maximize mutual information
between LiDAR reflectivity and image
intensity~\cite{pandey2015automatic}, while geometric approaches
align depth discontinuities with image
edges~\cite{levinson2013automatic,yuan2021pixel,ye2024mfcalib}.
Motion-based methods estimate extrinsics from ego-motion consistency
across sensors, requiring trajectories that excite all rotational
axes~\cite{taylor2016motion,persic2021spatiotemporal,lv2022observability}.
Learning-based regressors and matching
models~\cite{schneider2017regnet,iyer2018calibnet,lv2021lccnet,cattaneo2025cmrnext}
amortize calibration into a network, but their accuracy \rt{depends on}
the sensor configuration and scene statistics seen during training.
Across all three families, calibration accuracy depends on
whatever structure, motion, or appearance the environment provides.
\rt{Target-based calibration removes this dependency and remains the
standard in laboratory and production
settings}~\cite{zheng2025fastcalib,beltran2022automatic}.

\subsection{Target-Based Calibration}
\label{sec:rw_target}

Target-based methods differ mainly in the geometric primitive extracted
from the LiDAR data.
Planar checkerboards yield plane and edge
constraints~\cite{zhang2004extrinsic,zhou2018automatic},
but a single board plane constrains only its normal direction,
so multiple board poses are required to determine all six extrinsic
parameters.
Spherical targets provide a projection-invariant
center~\cite{toth2020automatic} at the cost of a single
correspondence per \rt{view}.
Circular holes cut into a planar board were introduced to obtain
four well-defined point correspondences from one
view~\cite{guindel2017automatic,beltran2022automatic}, \rt{which fix the
full transform without a second board pose}.
This design has since been extended with fiducial markers for
camera-side pose
recovery~\cite{domhof2019extrinsic,zheng2025fastcalib} and
combined with checkerboards for joint intrinsic--extrinsic
estimation~\cite{yan2023joint}.
Orthogonal to the choice of target, globally optimal solvers
remove the dependence on
initialization~\rt{\cite{jiao2023lce,zhu2025pointcloud,hu2025amalign}}; however, a
global optimum of an unconstrained feature model still inherits the
bias of that model, which the two priors remove.

Since the calibration board is manufactured to specification, its
dimensions are known exactly, and how a method exploits that
knowledge distinguishes the present work from existing
four-hole pipelines.
In prior work, the known hole radius enters as a tolerance range
that bounds the RANSAC circle model during
fitting~\cite{guindel2017automatic,beltran2022automatic,domhof2019extrinsic}:
the radius remains a free parameter within that range, so the
center--radius degeneracy under sparse angular coverage is left
intact.
The known rectangular layout is likewise \rt{applied} after estimation:
as a combinatorial check that selects mutually consistent
candidates~\cite{guindel2017automatic,beltran2022automatic}, or
as an optional rigid template alignment applied to unconstrained
circle-fit
results~\cite{domhof2019extrinsic,domhof2021joint}.
Because the template alignment operates on centers from
unconstrained fits, it cannot recover a hole whose fit was already
degenerate.
Concurrently, Jiang et
al.~\cite{jiang2025circular} jointly recover center, normal, and
radius with a conformal-geometric-algebra estimator, but the
radius is estimated rather than fixed and each circle is \rt{estimated}
independently, leaving both priors unused.

Prior work bounds the radius within a tolerance; we fix it to the
CAD value and estimate only the center together with a single
shared mixed-pixel bias, so that the degenerate direction is
removed from the fit.
The rectangular layout is then enforced as a global rigidity
constraint on centers that are already radius-constrained,
accepted only when its residual indicates agreement with the
observation.

\section{Problem Formulation}
\label{sec:pre}

This section formalizes the four-hole calibration pipeline
(Section~\ref{sec:formulation}) and analyzes the stage that determines
the accuracy of the recovered extrinsic
(Section~\ref{sec:bottleneck}).

\subsection{Four-Hole Calibration Pipeline}
\label{sec:formulation}

Let $\fL$, $\fC$, and $\fB$ denote the LiDAR, camera, and board
frames.
A subscript pair specifies the direction of a transform:
$\bm{R}_{\fC\fL}$ and $\bm{t}_{\fC\fL}$ map from $\fL$ to $\fC$,
so that $\bm{p}^{\fC} = \bm{R}_{\fC\fL}\,\bm{p}^{\fL} +
\bm{t}_{\fC\fL}$, with $\bm{R}_{\fC\fL} \in \mathrm{SO}(3)$ and
$\bm{t}_{\fC\fL} \in \mathbb{R}^3$.
A hat denotes an estimate and a star denotes an optimum.
Indices $m = 1,\dots,M$ and $k = 1,\dots,4$ enumerate scenes and
holes; $m$ is omitted when a single scene is considered.

The pipeline follows the two branches of Fig.~\ref{fig:overview}: the
camera branch recovers the board pose from the markers, the LiDAR branch
extracts the hole centers from the board plane, and a closed-form
registration combines them.
The calibration board \rt{has} four circular holes of nominal radius
$r$ whose centers form a $w \times h$ rectangle, together with four
ArUco markers~\cite{garrido2014automatic,guindel2017automatic,beltran2022automatic,zheng2025fastcalib}.
Both the radius and the layout are specified by the CAD model, so the
design centers $\bm{c}^{\fB}_k = (\pm w/2,\, \pm h/2)^{\top}$,
$k = 1,\dots,4$, are known exactly in the board plane.
On the image, marker corners observed at pixels $\bm{u}_j$ have
known board coordinates $\bm{q}^{\fB}_j$, and the board pose is
obtained by minimizing the reprojection error
\begin{equation}
\label{eq:campose}
(\bm{R}^{\star}_{\fC\fB},\, \bm{t}^{\star}_{\fC\fB}) =
\argmin_{\bm{R},\, \bm{t}} \sum_{j}
\big\| \pi(\bm{K}, \bm{d},\, \bm{R}\,\bm{q}^{\fB}_j + \bm{t})
       - \bm{u}_j \big\|^2,
\end{equation}
where $\bm{K}$ and $\bm{d}$ are the intrinsic and distortion
parameters and $\pi(\cdot)$ is the projection function.
The camera-frame hole centers follow from the CAD layout as
$\bm{p}^{\fC}_k = \bm{R}^{\star}_{\fC\fB}\,
[(\bm{c}^{\fB}_k)^{\top}, 0]^{\top} + \bm{t}^{\star}_{\fC\fB}$,
requiring no circle fitting on the image.

In the point cloud, an ROI filter and RANSAC~\cite{fischler1981ransac}
extract the board plane with unit normal $\bm{n}$, yielding the inlier
set $\mathcal{P}$.
A rotation $\bm{R}_{\fB\fL}$ aligns $\bm{n}$ with the $z$-axis and
maps each inlier to board-plane coordinates
$\bm{x}_i \in \mathbb{R}^2$ with intensity $I_i$ and mean plane
height $\bar{z}$.
The four in-plane hole centers $\bm{c}_k \in \mathbb{R}^2$ are
estimated from $\{\bm{x}_i, I_i\}$ and lifted back to the LiDAR
frame as
\begin{equation}
\label{eq:backproject}
\bm{p}^{\fL}_k = \bm{R}^{\top}_{\fB\fL}\,
(\bm{c}^{\top}_k,\, \bar{z})^{\top}.
\end{equation}
The two branches yield four correspondences
$\bm{p}^{\fL}_k \leftrightarrow \bm{p}^{\fC}_k$ per scene.
For $M$ jointly calibrated scenes, the extrinsic is the closed-form
absolute-orientation solution~\cite{umeyama1991least}
\begin{equation}
\label{eq:extrinsic}
(\bm{R}^{\star}_{\fC\fL},\, \bm{t}^{\star}_{\fC\fL}) =
\argmin_{\bm{R} \in \mathrm{SO}(3),\, \bm{t} \in \mathbb{R}^3}
\sum_{m=1}^{M} \sum_{k=1}^{4}
\big\| \bm{R}\,\bm{p}^{\fL}_{mk} + \bm{t}
       - \bm{p}^{\fC}_{mk} \big\|^2,
\end{equation}
obtained from the SVD of the centered cross-covariance $\bm{\Sigma}$.

\subsection{Accuracy Analysis}
\label{sec:bottleneck}

The three stages of the pipeline contribute unequally to the final
extrinsic error.
Equation~\eqref{eq:extrinsic} is closed-form and introduces no
error of its own; it only propagates the error in its inputs.
Plane segmentation aggregates every board point into three
parameters, so individual range noise averages out and the stage is
well conditioned.
The camera-frame centers $\bm{p}^{\fC}_k$ are generated by one
shared board pose~\eqref{eq:campose}, so they cannot deform relative
to one another: an error in the board pose displaces all four centers
consistently and is largely absorbed by the extrinsic.
\rt{The LiDAR centers $\bm{c}_k$ have neither property.}
Each is recovered from only those points that fall on one hole
boundary, and each is estimated independently of the other three.
The resulting errors are uncorrelated and enter~\eqref{eq:extrinsic}
as correspondence noise.
Calibration accuracy is therefore governed by this single stage,
which is also where the board geometry is available but, as
Section~\ref{sec:rw} discussed, left unused.
This observation motivates the pattern-prior formulation developed
in Section~\ref{sec:method}.

\section{Pattern-Prior Extrinsic  Calibration}
\label{sec:method}

This section develops the two geometric priors that address the
fitting degeneracy and the lack of inter-hole coupling identified in
Section~\ref{sec:bottleneck}, as illustrated in Fig.~\ref{fig:priors}.

\subsection{Free-Circle Ambiguity}
\label{sec:degeneracy}

\begin{figure}[!t]
\centering
\includegraphics[width=0.9\linewidth]{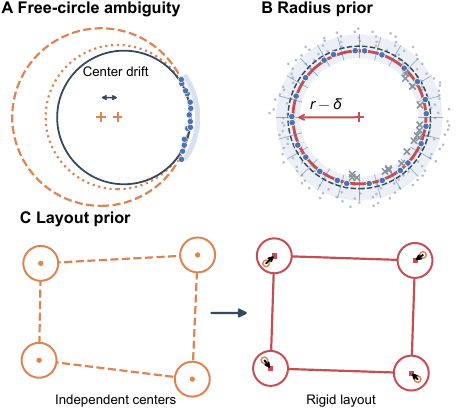}
\caption{Two pattern priors. (\textbf{A}) A short measured arc (blue)
permits multiple center--radius pairs (orange); dark curve: reference
circle. (\textbf{B}) Sector samples constrain the center at radius
$r-\delta$; crosses mark rejected returns. (\textbf{C}) A rigid CAD layout
couples the four independent centers (orange), producing the projected
centers (red).}
\label{fig:priors}
\end{figure}

Let $\bar{\bm{v}}_k$ be the normalized mean direction of the boundary
samples of hole $k$ seen from its center, and let $\phi_{ik}$ be the
polar angle of sample $i$ about that direction, so that the samples
occupy an angular span
$\Phi_k = \max_i \phi_{ik} - \min_i \phi_{ik}$.
Under the free-circle parameterization, in which both the center
$\bm{c}_k$ and the radius $r_k$ are unknown, consider the joint
perturbation that translates the center by $\Delta\bar{\bm{v}}_k$
and reduces the radius by $\Delta$.
With $d_{ik} = \|\bm{x}_i - \bm{c}_k\|$, the distance to the
displaced center is
$d_{ik} - \Delta\cos\phi_{ik} + O(\Delta^2)$, so the residual
changes by
\begin{equation}
\label{eq:degeneracy}
\Delta e_{ik} = \Delta\big(1 - \cos\phi_{ik}\big) + O(\Delta^2).
\end{equation}
The change is first order in $\Delta$ but quadratic in the angular
deviation.
For samples confined to $|\phi_{ik}| \le \Phi_k/2$ it is bounded by
$\Delta\Phi_k^2/8$, so the squared residual perturbation scales as
$O(\Delta^2\Phi_k^4)$, flattening as the fourth power of the angular
span~\cite{alsharadqah2009error}.
As shown in Fig.~\ref{fig:priors}A, a short measured arc admits
multiple center--radius pairs whose residuals differ by
$O(\Delta^2\Phi_k^4)$, below the range noise for the spans observed here.
This ambiguity has two effects.
The center is weakly determined along $\bar{\bm{v}}_k$ even when the
range measurements are accurate, and mixed-pixel
returns~\cite{tuley2005artifacts}, which
displace the observed boundary inward along the same direction,
are absorbed by the estimator as an apparent reduction of radius.
Fixing the radius to the CAD value
removes~\eqref{eq:degeneracy} from the parameter space at no
modeling cost, since the board is manufactured to specification.

\subsection{Radius Prior}
\label{sec:radius_prior}

Mixed pixels reduce the observed radius of every hole by a comparable
amount, as the effect is governed by the beam footprint at a given range.
\rt{We model this shrinkage with a single bias
$\delta \in [0, \delta_{\max}]$ shared by all four holes, and fit each
center against the effective radius $r - \delta$}
(Fig.~\ref{fig:priors}B).
The residual of point $i$ on hole $k$ is
\begin{equation}
\label{eq:fixed_radius}
e_{ik} = d_{ik} - (r - \delta),
\end{equation}
reducing the unknowns of each hole from three to two and adding one
scalar for the whole board.

\rt{We draw candidates for hole $k$ from the annulus}
$\mathcal{A}_k = \{i : r{-}\delta{-}b_{\mathrm{in}} \le d_{ik} \le
r{-}\delta{+}b_{\mathrm{out}}\}$ \rt{and retain one boundary
representative per azimuth sector}, the innermost surviving point of
that sector, which \rt{prevents} a densely sampled arc from dominating the
fit. \rt{We update the hole only when the retained samples
$\mathcal{E}_k$ occupy} at least ten of the $S = 24$ sectors with no
empty run longer than $S/4$.

The center of hole $k$ minimizes
$\sum_{i \in \mathcal{E}_k} \ell_\kappa(e_{ik})$ under the Huber
loss~\cite{huber1964robust} $\ell_\kappa$, \rt{whose threshold we set to}
$\kappa = \SI{12}{mm}$.
With $\bm{v}_{ik} = (\bm{x}_i - \bm{c}_k)/d_{ik}$ and weights
$w_{ik} = \min(1, \kappa/|e_{ik}|)$, each Gauss--Newton step solves
\begin{equation}
\label{eq:irls}
\bm{H}_k \Delta\bm{c}_k = \bm{g}_k, \qquad
\bm{c}_k \leftarrow \bm{c}_k + \Delta\bm{c}_k,
\end{equation}
where $\bm{H}_k = \sum_i w_{ik}\, \bm{v}_{ik}\bm{v}_{ik}^\top$ and
$\bm{g}_k = \sum_i w_{ik}\, e_{ik}\, \bm{v}_{ik}$.
After updating every valid hole, the shared bias is refreshed from
the weighted mean observed radii
$\hat{r}_k = W_k^{-1}\sum_i w_{ik}\, d_{ik}$
($W_k = \sum_i w_{ik}$):
\begin{equation}
\label{eq:bias_update}
\delta \leftarrow
\min\!\Big(\delta_{\max},\;
\max\!\big(0,\;
r - |\mathcal{V}|^{-1}{\textstyle\sum_{k \in \mathcal{V}}}
\hat{r}_k
\big)\Big),
\end{equation}
with $\delta_{\max} = \SI{30}{mm}$.
Sharing one bias across holes avoids reintroducing four free radii;
its identifiability depends on the diversity of boundary
directions, as Section~\ref{sec:theory_joint} analyzes.

\subsection{Layout Prior}
\label{sec:rigid_rect}

Under the radius prior alone, the four holes remain independently
parameterized, and their eight center coordinates are uncoupled.
The layout prior removes this independence by constraining the four
centers to a rigid placement of the CAD rectangle
(Fig.~\ref{fig:priors}C),
$\bm{c}_k = \bm{R}(\theta)\,\bm{c}^{\mathcal{B}}_k + \bm{t}$
with $\bm{R}(\theta) \in \mathrm{SO}(2)$ and
$\bm{t} \in \mathbb{R}^2$, collapsing the eight coordinates to
three continuous degrees of freedom and projecting out five
deformation directions.
Given the radius-constrained estimates $\{\hat{\bm{c}}_k\}$, the
optimal rigid placement and corner assignment are
\begin{equation}
\label{eq:rigid_align}
(\theta^\star, \bm{t}^\star, \sigma^\star) =
\argmin_{\theta,\, \bm{t},\, \sigma} \sum_{k=1}^{4}
\big\| \hat{\bm{c}}_k - \bm{R}(\theta)\,
\bm{c}^{\mathcal{B}}_{\sigma(k)} - \bm{t} \big\|^2,
\end{equation}
where $\sigma$ is a permutation assigning CAD corners to detected
centers.
For a fixed $\sigma$ this is a planar Procrustes problem.
With centered vectors
$\bm{a}_k = \bm{c}^{\mathcal{B}}_{\sigma(k)} -
\bar{\bm{c}}^{\mathcal{B}}$
and $\bm{b}_k = \hat{\bm{c}}_k - \bar{\bm{c}}$, the closed-form
solution is
\begin{equation}
\label{eq:theta_star}
\begin{gathered}
\theta^\star = \operatorname{atan2}\!\bigg(
  \sum_{k} \bm{a}_k \times \bm{b}_k,\;
  \sum_{k} \bm{a}_k \cdot \bm{b}_k \bigg), \\[2pt]
\bm{t}^\star = \bar{\bm{c}} -
  \bm{R}(\theta^\star)\,\bar{\bm{c}}^{\mathcal{B}},
\end{gathered}
\end{equation}
where $\bm{a}_k \times \bm{b}_k = a_{k,x}b_{k,y} - a_{k,y}b_{k,x}$
and the bars denote four-center means.
\rt{We enumerate all $4! = 24$ assignments}; the relabeling symmetry of
the rectangle resolves the mirror ambiguity implicitly.
Because a rigid fit distributes a single gross error across all four
centers, \rt{we apply the projection} only when every hole passed the
coverage test and the projected centers satisfy
\begin{equation}
\label{eq:accept}
\max_{1 \le k \le 4}
\|\hat{\bm{c}}_k - \bm{c}^\star_k\| \le d_{\max},
\end{equation}
with $d_{\max} = \SI{35}{mm}$; otherwise \rt{we retain the fits of each
hole}.

\begin{algorithm}[t]
\caption{Pattern-Prior Extrinsic Calibration}
\label{alg:p2calib}
\begin{algorithmic}[1]
\Require Board pts $\{\bm{x}_i, I_i\}_{i \in \mathcal{P}}$, CAD
         layout $\{\bm{c}^{\mathcal{B}}_k\}$, radius $r$
\Ensure  Extrinsic $(\bm{R}^{\star}_{\mathcal{CL}},\, \bm{t}^{\star}_{\mathcal{CL}})$
\State Initialize $\{\bm{c}_k, \hat r_k\}$ from the plane-raster
       circle candidates and $\delta$ by~\eqref{eq:bias_update}
\Statex \emph{// Radius prior (Section~\ref{sec:radius_prior})}
\For{$n = 1$ to $N_{\mathrm{out}} = 5$}
  \For{$k = 1$ to $4$}
    \State Annulus selection, mixed-pixel rejection, sector sampling
    \If{coverage test passed}
      \For{$m = 1$ to $N_{\mathrm{in}} = 3$}
        \State Update $\bm{c}_k$ at fixed radius $r - \delta$ by~\eqref{eq:irls}
      \EndFor
    \EndIf
  \EndFor
  \State Update shared bias $\delta$ by~\eqref{eq:bias_update}
\EndFor
\Statex \emph{// Layout prior (Section~\ref{sec:rigid_rect})}
\State Solve rigid rectangular projection~\eqref{eq:rigid_align} via~\eqref{eq:theta_star}
\If{acceptance test~\eqref{eq:accept} passed}
  \State Snap $\bm{c}_k \leftarrow \bm{c}^\star_k$ for all $k$
\EndIf
\Statex \emph{// Registration}
\State Lift $\bm{c}_k$ to $\bm{p}^{\mathcal{L}}_k$ by~\eqref{eq:backproject}
\State Solve extrinsic~\eqref{eq:extrinsic}
\end{algorithmic}
\end{algorithm}

\subsection{Estimation Model and Solver}
\label{sec:joint}

The two priors together describe the board by three pose parameters
$(\theta, \bm{t})$ and one shared bias $\delta$, estimated from the
full set of boundary samples:
\begin{equation}
\label{eq:joint_obj}
\min_{\theta,\, \bm{t},\, \delta}\;
\sum_{k=1}^{4} \sum_{i \in \mathcal{E}_k}
\ell_\kappa\!\Big(
\big\| \bm{x}_i - \bm{R}(\theta)\,\bm{c}^{\mathcal{B}}_k
       - \bm{t} \big\| - (r - \delta)\Big).
\end{equation}
Equation~\eqref{eq:joint_obj} is the estimation model whose
parameter count Section~\ref{sec:theory_dof} analyzes.
Because the sector assignment depends on the working centers, a
joint minimization would re-select its own samples at every step.
The solver therefore minimizes a nine-parameter relaxation, in which
the four centers move freely under the shared bias, and projects
the result onto the rigid rectangle
by~\eqref{eq:rigid_align}--\eqref{eq:theta_star} under the
acceptance test~\eqref{eq:accept}.
Algorithm~\ref{alg:p2calib} summarizes the full process, and
Section~\ref{sec:theory} analyzes the properties of the two priors.

\section{Analysis of the Pattern Priors}
\label{sec:theory}

This section establishes why the two priors of
Section~\ref{sec:method} restore a well-posed estimate under sparse
angular coverage.
Section~\ref{sec:theory_dof} counts the parameters they remove,
Sections~\ref{sec:theory_rp} and~\ref{sec:theory_lp} quantify the
conditioning and error effects, and
Section~\ref{sec:theory_joint} shows that the two priors are
mutually dependent.

\subsection{Degrees of Freedom}
\label{sec:theory_dof}

A free circle \rt{has} three continuous unknowns (an in-plane center
and a radius), so the free-circle parameterization of the four holes
has twelve.
The radius prior (RP) replaces the four independent radii by the
known $r$ acted on by a single shared bias $\delta$, and the layout
prior (LP) collapses the eight center coordinates onto the rigid
pose $(\theta, \bm{t})$ of the CAD rectangle:
\begin{equation}
\label{eq:dof_chain}
\underbrace{4 \times 3}_{12}
\;\xrightarrow{\ \mathrm{RP}\ }\;
\underbrace{4 \times 2 + 1}_{9}
\;\xrightarrow{\ \mathrm{LP}\ }\;
\underbrace{3 + 1}_{4}.
\end{equation}
The number of measurements is unchanged by this reduction.
Each hole contributes at most $S$ sector samples, so \rt{one view} offers
on the order of $4S$ boundary measurements whichever
parameterization consumes them, and the measurement-to-parameter
ratio rises from $4S/12$ to $4S/4$ without discarding a single
measurement.
\rt{The reduction removes only parameter directions the measurements
cannot resolve}, the radius direction identified
by~\eqref{eq:degeneracy} and the five deformation directions
identified by the rectangle rigidity.

\subsection{Conditioning under the Radius Prior}
\label{sec:theory_rp}

Consider one hole under the free-circle parameterization, with
unknowns $(\bm{c}_k, r_k)$ and residual $d_{ik} - r_k$.
The Jacobian row of sample $i$ is $(-\bm{v}^{\top}_{ik},\, -1)$,
giving the normal matrix
\begin{equation}
\label{eq:hfull}
\tilde{\bm{H}}_k =
\begin{bmatrix}
\bm{H}_k & \bm{s}_k \\
\bm{s}^{\top}_k & W_k
\end{bmatrix},
\qquad
\bm{s}_k = \sum_{i \in \mathcal{E}_k} w_{ik}\, \bm{v}_{ik},
\end{equation}
with $\bm{H}_k$ the scatter matrix of~\eqref{eq:irls}.
\rt{Conditioning of such a normal matrix is commonly examined by
decoupling its blocks~\cite{hu2026dcreg}; the coupling of interest here
is the one between the center and the radius.}
By~\eqref{eq:degeneracy}, the unit direction
$(\bar{\bm{v}}^{\top}_k,\, -1)^{\top}/\sqrt{2}$ produces a residual
change bounded by $\Phi_k^2/8$ for each sample, so
\begin{equation}
\label{eq:lam_full}
\lambda_{\min}(\tilde{\bm{H}}_k) = O(W_k \Phi_k^{4}).
\end{equation}
Fixing the radius deletes the last row and column
of~\eqref{eq:hfull}, leaving $\bm{H}_k$ alone.
Writing
$\bm{v}_{ik} = (\cos\phi_{ik},\, \sin\phi_{ik})^{\top}$, the
eigenvalues of $\bm{H}_k$ admit the exact form
\begin{equation*}
\begin{gathered}
\lambda_{\min,\max}(\bm{H}_k) = \tfrac{1}{2} W_k (1 \mp R_k), \\[2pt]
R_k = W_k^{-1} \Big\| \textstyle\sum_{i \in \mathcal{E}_k} w_{ik}\,
e^{\mathrm{j}\, 2\phi_{ik}} \Big\| \in [0,\, 1],
\end{gathered}
\end{equation*}
where the second-order circular resultant $R_k$ measures the directional
concentration of the retained boundary samples.
For uniform sampling over a span $\Phi_k \le \pi$,
$R_k = \sin\Phi_k / \Phi_k$ and
$\lambda_{\min}(\bm{H}_k) \approx W_k \Phi_k^{2}/12$.
Comparing with~\eqref{eq:lam_full},
\begin{equation}
\label{eq:cond_gain}
\mathrm{cond}(\tilde{\bm{H}}_k) = O(\Phi_k^{-4}),
\qquad
\mathrm{cond}(\bm{H}_k) = \frac{1+R_k}{1-R_k} = O(\Phi_k^{-2}),
\end{equation}
so the radius prior reduces the rate of conditioning loss from
$O(\Phi_k^{-4})$ to $O(\Phi_k^{-2})$ as the angular span narrows.
The resultant $R_k$ also detects a second failure mode invisible to
the span alone: samples concentrated at two diametrically opposite
positions give $R_k = 1$ and a singular $\bm{H}_k$, which is why
the coverage test preceding~\eqref{eq:irls} constrains both the
sector count $N_k$ and the longest empty run $L_k$.

\subsection{Error Projection under the Layout Prior}
\label{sec:theory_lp}

Define the concatenated center vector
$\hat{\bm{c}} = (\hat{\bm{c}}^{\top}_1, \dots,
\hat{\bm{c}}^{\top}_4)^{\top} \in \mathbb{R}^{8}$.
For a fixed assignment $\sigma$, the rigid placements of the CAD
rectangle trace a three-dimensional manifold
$\mathcal{M}_\sigma \subset \mathbb{R}^{8}$ generated by
$(\theta, \bm{t})$, and~\eqref{eq:rigid_align} returns the closest
point on $\bigcup_\sigma \mathcal{M}_\sigma$, with the closed form
of~\eqref{eq:theta_star} realizing the orthogonal projection for the
selected assignment.

Let $\hat{\bm{c}}_k = \bm{c}^{\mathrm{gt}}_k + \bm{\eta}_k$ for the
true rigid placement $\bm{c}^{\mathrm{gt}}_k$, and model
$\bm{\eta}_k$ as independent, zero-mean, and isotropic with variance
$s^2$ in each coordinate.
The tangent space of $\mathcal{M}_\sigma$ at
$\bm{c}^{\mathrm{gt}}$ is exactly three-dimensional, so to first
order
\begin{equation}
\label{eq:proj_reduction}
\mathbb{E}\|\hat{\bm{c}} - \bm{c}^{\mathrm{gt}}\|^2 = 8s^2,
\qquad
\mathbb{E}\|\bm{c}^\star - \bm{c}^{\mathrm{gt}}\|^2 \simeq 3s^2,
\end{equation}
since orthogonal projection onto a $d$-dimensional subspace of
$\mathbb{R}^n$ retains expected squared norm $ds^2$ from an
isotropic input.
Under this model and a correct assignment, the layout prior retains
$3/8$ of the expected squared center error, equivalently
$\sqrt{3/8} \approx 0.61$ in root-mean-square.
The isotropy assumption is an idealization, since the radius prior
constrains the radial direction more tightly than the tangential one;
$3/8$ is accordingly a limiting value.
The discarded five-dimensional component corresponds to the family of
rectangle deformations no rigid board can produce, and its magnitude
is what the acceptance test~\eqref{eq:accept} thresholds.
A component large enough to fail that test indicates an unreliable
hole estimation.

\subsection{Complementarity of the Two Priors}
\label{sec:theory_joint}

\rt{Each prior is necessary for the other to be well posed.}
The dependence of LP on RP is made explicit in
Section~\ref{sec:rigid_rect}: projection redistributes the residual
across all four centers, so an input center whose own fit was
degenerate contaminates the other three.
RP bounds that input error through~\eqref{eq:cond_gain}.

The reverse dependence concerns the identifiability of $\delta$.
Under the fully constrained parameterization the Jacobian column of
$\delta$ is constant while that of $\bm{t}$ is $-\bm{v}^{\top}_{ik}$, so
$\delta$ is unidentifiable whenever some unit $\bm{u}$ satisfies
$\bm{v}_{ik} \cdot \bm{u} = \mathrm{const}$ over every retained sample,
which is what a single hole on a narrow arc gives: there
$\bm{v}_{ik} \approx \bar{\bm{v}}_k$ and $\delta$ is indistinguishable
from a translation along $\bar{\bm{v}}_k$.
The rectangular layout breaks this condition, because the four holes
present their boundary directions in four distinct global
orientations, and \rt{the rigidity of the layout prevents a rectangle
free to scale from trading its scale factor against $\delta$}.
Together,~\eqref{eq:cond_gain} and~\eqref{eq:proj_reduction} account
for the behavior of~\eqref{eq:joint_obj} under sparse angular
coverage: \rt{RP bounds the normal matrix of every hole away from
singularity}, LP removes the coordinate freedom that coverage alone
cannot constrain and renders the shared bias identifiable, and the
four surviving parameters are estimated from the full set of boundary
samples.
Section~\ref{sec:exp_coverage} probes this analysis directly by
varying boundary coverage and standoff distance.

\section{System Implementation}
\label{sec:impl}

\rt{We build} P$^2$Calib on the FAST-Calib\footnote{\url{https://github.com/hku-mars/FAST-Calib}}~\cite{zheng2025fastcalib}
codebase \rt{and integrate} the radius prior and layout prior of
Sections~\ref{sec:radius_prior} and \ref{sec:rigid_rect} into an
interactive calibration tool.
\rt{We implement it} in Python~3.10 with Open3D~0.19 for point
cloud processing and OpenCV for image processing.
\rt{For the timings reported in Section~\ref{sec:exp_runtime} we run the
tool} on a desktop with an Intel i7-12700K processor using a single
thread.
The tool reports the intermediate result of each pipeline stage and
reports both the layout disagreement~\eqref{eq:accept} and the
registration residual~\eqref{eq:extrinsic} for every \rt{view}, so that
the \rt{practitioner} can distinguish extraction quality from correspondence
quality and can exclude a \rt{view} before the multi-scene solve.
\rt{We designed the annulus refinement, including the shared bias
$\delta$ and the mixed-pixel rejection, for} sensors whose hole
boundaries are eroded by mixed pixels, such as the area-array LiDAR
evaluated in Section~\ref{sec:exp_real}.
On scanning LiDARs such as the Livox Avia and Mid-360, the raster-based
extraction \rt{provides} dense hole candidates. The radius prior then
fixes every radius to $r$ without $\delta$, and the layout prior acts on
the raster centers.

\begin{figure}[t]
\centering
\includegraphics[width=\linewidth]{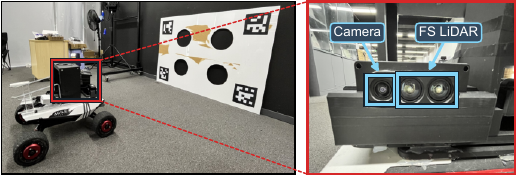}\\[2.5pt]
\includegraphics[width=\linewidth]{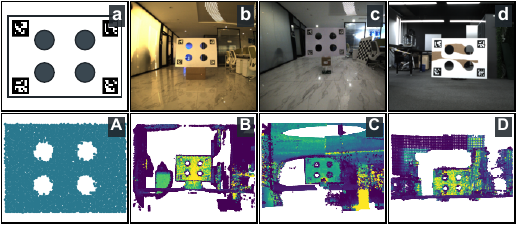}
\caption{\rt{Experiment setup.} \textbf{Top}: the mobile platform and
calibration board, with the camera and LiDAR enlarged. \textbf{Bottom}:
one data source per column, camera image (\textbf{a--d}) above the
range-shaded LiDAR scan (\textbf{A--D}), the holes appearing as white
voids; (\textbf{a,A}) simulator, (\textbf{b,B}) Avia, (\textbf{c,C})
Mid-360, (\textbf{d,D}) the area-array FS sensor.}
\label{fig:setup}
\end{figure}

\section{Experiments}
\label{sec:exp}

We validate the two pattern priors on simulated boards with known
ground truth (Section~\ref{sec:exp_syn}), which also tests the
predictions of Section~\ref{sec:theory}, and on real sensor data across
multiple LiDAR types (Section~\ref{sec:exp_real}).

\begin{figure}[t]
\centering
\includegraphics[width=0.95\linewidth]{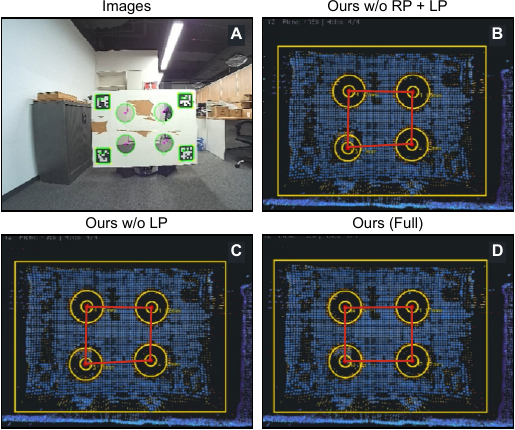}
\caption{Real-scene ablation in the FS-B dataset. (\textbf{A}) Camera
detections; (\textbf{B}) both priors off: fitted radii vary across holes; (\textbf{C}) radius prior only: centers form a skewed quadrilateral;
(\textbf{D}) both priors: centers conform to the rectangular layout. Red lines join the four detected centers; blue
cloud marks are thickened for legibility.}
\label{fig:ablation_vis}
\vspace{-0.1cm}
\end{figure}

\subsection{Experimental Setup}
\label{sec:setup}

\textbf{Data.}
\rt{We generate the simulated boards with a four-hole simulator.} It reproduces the CAD model of Section~\ref{sec:formulation}, a $1.40 \times 1.00$\,m plate with
four $r = 0.12$\,m holes on a $0.50 \times 0.40$\,m rectangle, and adds
Gaussian range noise along the ray ($\sigma = 12$--$14$\,mm, the robust
plane-fit residual measured on the real FS boards), 2\,mm quantization,
and mixed-pixel returns intruding up to 20\,mm into each hole.
The acquisition protocol mirrors a real session: six standoff distances
grouped as near at 1.5 and 2.0\,m, mid at 2.5 and 3.2\,m and far at 4.0
and 5.0\,m; five placements per distance, left, center and right at
mid-height and up and down on the center column, each on an image quarter
point and clamped to both fields of view; and two frames per placement,
60 frames in all.
A random tilt of up to $\pm 5^\circ$ sets the lateral placements to
$37.5^\circ$ incidence and the vertical ones at $8^\circ$--$16^\circ$.
A wall 1\,m behind the board, excluded by the \rt{practitioner} ROI but kept by
velo2cam as its edge test requires, completes every scene.
Each frame is rendered at two sampling densities, a single frame (SF)
at \rt{the $0.3^\circ$ pitch of the sensor}, 19.6\,mm at 3.2\,m and scaled with
range, and an accumulated cloud (AC) at 5\,mm; geometry and noise are fixed per frame, and the three detector
seeds, which vary only the plane RANSAC, give identical inlier sets, so
the reported values have zero spread across seeds.

For the real-world evaluation, we collect two datasets (FS-B and
FS-C) using a mobile platform equipped with an area-array
solid-state LiDAR and a $1920 \times 1080$ camera, as shown in
Fig.~\ref{fig:setup}.
The area-array LiDAR has a $120^\circ \times 50^\circ$ field of view,
$0.33^\circ$ angular resolution in both axes, a range of 0.2--48\,m,
and a range noise of $\pm$50\,mm.
FS-B contains 18 scenes at 1.5--4.3\,m and FS-C contains 20 scenes
at 1.0--4.2\,m, both with the same board geometry and fixed
camera--LiDAR mounting.
To evaluate cross-sensor generality, \rt{we also adopt} three
scanning-LiDAR datasets provided by
FAST-Calib~\cite{zheng2025fastcalib}: one Livox Avia session (5
scenes, solid-state with non-repetitive scan), and two Livox Mid-360
sessions (4 and 3 scenes, hybrid solid-state with $360^\circ$
horizontal coverage).
Figure~\ref{fig:setup} shows representative inputs from each
sensor.

\textbf{Baselines.}
RP and LP denote the radius prior and the layout prior throughout.
\rt{We give every variant the same camera processing and SVD registration,
and compare against}
\begin{itemize}
    \item \textbf{velo2cam}\footnote{\url{https://github.com/beltransen/velo2cam_calibration}}~\cite{beltran2022automatic}:
    the original four-hole detector. Its ring-wise edge test assumes
    organized scan rings and the background seen through the holes,
    which solid-state clouds do not provide, so we adapt its input by
    resampling the cloud into 128 elevation-binned virtual rings and
    raising the range threshold from 6 to 15\,m; every detector
    parameter retains its default value.
    \item \textbf{FAST-Calib}\footnote{\url{https://github.com/hku-mars/FAST-Calib}}~\cite{zheng2025fastcalib}:
    the Livox adaptation of the four-hole board, which extracts the
    circles from a rasterization of the board plane. It is our primary
    baseline and differs from P$^2$Calib only in lacking both priors.

    \item \textbf{Ours w/o LP} and \textbf{Ours w/o RP}: the radius
    prior and the layout prior in isolation.
\end{itemize}
\textbf{Metrics.}
In simulation the simulator provides the true hole centers and
extrinsic. The first metric is the mean hole-center error [mm] of each
frame, which counts as detected when four centers lie within 50\,mm of
the truth. \rt{We then fit the extrinsic jointly on all detected frames and score
it by} its rotation [$^\circ$] and translation [mm] error against
the truth, together with the registration residual
of~\eqref{eq:extrinsic}, a fitting residual \rt{we report} for completeness.
The last metric is a leave-one-out (LOO) reprojection error: \rt{we fit
the extrinsic on the remaining frames, then project the centers of the
held-out frame and compare them with} the true pixel positions.
On real data, without ground truth, the joint residual and the LOO error
are measured against the detected ellipse centers, \rt{as in map-quality
assessment without a dense reference}~\rt{\cite{hu2025mapeval}}.

\begin{figure}[t]
\centering
\includegraphics[width=\linewidth]{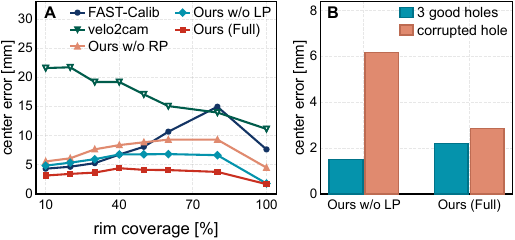}
\caption{Rim-coverage study on AC boards at 3.2\,m. (\textbf{A}):
hole-center error as the covered fraction of each rim drops from 100\% to
10\%; \rt{each method is scored on its own successful detections}, and hollow
markers denote partial success. (\textbf{B}): error of the three fully covered holes
(cyan) and of the hole reduced to 25\% coverage (salmon).}
\label{fig:coverage}
\vspace{-0.1cm}
\end{figure}

\begin{figure*}[t]
\centering
\includegraphics[width=\textwidth]{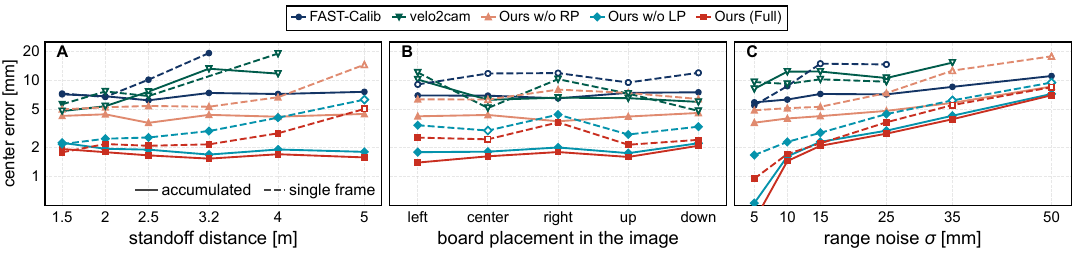}
\caption{Hole-center error on the simulated protocol against
(\textbf{A}) standoff distance, (\textbf{B}) board placement in the image
and (\textbf{C}) range-noise $\sigma$ on the mid group, on a shared
logarithmic scale; AC solid, SF dashed. A curve ends where the method
recovers no board.}
\label{fig:sweeps}
\vspace{-0.1cm}
\end{figure*}

\subsection{Simulated Experiments}
\label{sec:exp_syn}

\subsubsection{Overall results analysis}
\label{sec:exp_ablation}

Table~\ref{tab:ablation} reports the hole-center error of every variant
by standoff group, and Table~\ref{tab:synth_extrinsic} the joint
extrinsic fitted on all detected frames.
Figure~\ref{fig:ablation_vis} shows the progression on one real \rt{scene}:
independently fitted circles, radius-constrained circles, and the full
pipeline with the layout projection.
Where FAST-Calib recovers the board, the full pipeline reduces the
hole-center error from 7.0--14.7\,mm to 2.0--2.1\,mm on SF and from
6.8--7.4\,mm to 1.6--1.9\,mm on AC frames, a $3.5$--$6.9\times$
reduction. On the far group of SF frames
FAST-Calib recovers no board at all, whereas the full pipeline detects
57 of 60 trials at 3.9\,mm.
The improvement propagates to the extrinsic: the LOO reprojection
error falls from 2.61 to 0.40\,px on SF and from 1.51 to 0.23\,px on
AC frames, and the rotation and translation errors of
Table~\ref{tab:synth_extrinsic} fall by 3.6 and 4.7 times on SF frames
and by 3.6 and 10 times on AC frames.
Restricting the fit to the 40 SF frames that every variant detects
leaves the ordering unchanged, at 0.36 against 2.61\,px. This is also
the set that FAST-Calib itself recovers.
velo2cam, run on the same frames with its background wall, detects 25 and
28 of 60 frames and reaches 2.28 and 1.95\,px.

\begin{table}[t]
\caption{Hole-center error [mm] $\downarrow$ in simulation by standoff group}
\label{tab:ablation}
\centering
\footnotesize
\renewcommand{\arraystretch}{1.45}
\begin{threeparttable}
\setlength{\tabcolsep}{4pt}
\begin{tabularx}{\linewidth}{|l||>{\centering\arraybackslash}X|>{\centering\arraybackslash}X|>{\centering\arraybackslash}X||>{\centering\arraybackslash}X|>{\centering\arraybackslash}X|>{\centering\arraybackslash}X|}
\hline
\rowcolor{tabhead}
 & \multicolumn{3}{c||}{\textbf{Single frame (SF)}} & \multicolumn{3}{c|}{\textbf{Accumulated (AC)}} \\
\cline{2-7}
\rowcolor{tabhead}
\multirow{-2}{*}{\textbf{Variant}} & Near & Mid & Far & Near & Mid & Far \\
\hline
velo2cam\tnote{a} & \cellcolor{tabbest!25}6.5 & \cellcolor{tabbest!25}6.9 & \cellcolor{tabbest!25}19.0 & \cellcolor{tabbest!25}5.1 & \cellcolor{tabbest!8}8.3 & \cellcolor{tabbest!8}11.8 \\
\hline
FAST-Calib & \cellcolor{tabbest!8}7.0 & \cellcolor{tabbest!8}14.7 & -- & \cellcolor{tabbest!8}7.0 & \cellcolor{tabbest!25}6.8 & \cellcolor{tabbest!25}7.4 \\
Ours w/o RP & \cellcolor{tabbest!45}5.1 & \cellcolor{tabbest!45}5.4 & \cellcolor{tabbest!45}10.4\tnote{b} & \cellcolor{tabbest!45}4.4 & \cellcolor{tabbest!45}4.0 & \cellcolor{tabbest!45}4.3 \\
Ours w/o LP & \cellcolor{tabbest!70}2.3 & \cellcolor{tabbest!70}2.8 & \cellcolor{tabbest!70}5.1\tnote{b} & \cellcolor{tabbest!70}2.1 & \cellcolor{tabbest!70}1.8 & \cellcolor{tabbest!70}1.9 \\
\hline
Ours (Full) & \cellcolor{tabbest}\textbf{2.0} & \cellcolor{tabbest}\textbf{2.1} & \cellcolor{tabbest}\textbf{3.9}\tnote{b} & \cellcolor{tabbest}\textbf{1.9} & \cellcolor{tabbest}\textbf{1.6} & \cellcolor{tabbest}\textbf{1.6} \\
\hline
\end{tabularx}
\begin{tablenotes}[flushleft]
\item \scriptsize\textbf{Note:} Mean over successful trials, 60 per group; a darker cell marks a lower error within a column, \textbf{bold} the best, \textquotedblleft--\textquotedblright{} no detection. \scriptsize Deterministic, one run per frame; detects 18/6/1 (SF) and 17/8/3 (AC) of 20 frames in the near/mid/far groups. \textsuperscript{b} Ours w/o RP, Ours w/o LP, Ours (Full): 57 of 60 trials (far, AC).
\end{tablenotes}
\end{threeparttable}
\end{table}

The two priors contribute in complementary regimes.
The radius prior is the larger contributor: alone it reaches
2.3--5.1\,mm on SF and 1.8--2.1\,mm on AC frames, 91 to 96\% of the
improvement of the full pipeline, which confirms that the center--radius
degeneracy of Section~\ref{sec:degeneracy}, and not the range noise,
dominates the baseline error.
The layout prior alone yields 4.0--10.4\,mm, and its contribution
concentrates where the angular coverage is sparsest: it lowers the
radius-only error of the far SF group from 5.1 to 3.9\,mm, whereas on
AC frames the radius prior alone is already within 0.3\,mm of the full
pipeline.
This behavior follows the design of Section~\ref{sec:rigid_rect}: the
layout constraint corrects the residual cross-hole errors that remain
after the fit of each hole, and those errors decrease as the angular
coverage increases.
Two controls on boards held at the center of the image, at 3.2\,m and
$5^\circ$ incidence, confirm the attribution. Giving FAST-Calib the
denser plane rasterization RP relies on, with neither prior, reaches
only 5.0--6.3\,mm. Under a violated prior, a radius error of
$\pm 1.5$\,mm on each hole with mixed-pixel intensities overlapping the
rejection threshold, the full pipeline still reaches 3.4\,mm against
7.3\,mm for the baseline.

\subsubsection{Coverage, standoff and noise analysis}
\label{sec:exp_coverage}

The rim-coverage sweep tests the conditioning analysis of
Section~\ref{sec:theory_rp} directly: it shortens the measured arcs on
an AC board at 3.2\,m while everything else is held fixed
(Fig.~\ref{fig:coverage}A).
As the covered fraction drops from 100\% to 10\%, the full pipeline
degrades from 1.7 to 3.2\,mm, while the baseline ranges from 4.3 to
15.0\,mm and is non-monotonic, because its error follows the mixed-pixel
corruption that survives the mask and not the number of retained
boundary samples;
velo2cam remains between 11 and 22\,mm, since the mask removes the edge
returns its gradient test requires.
The ablation curves separate the two priors: with full rims the radius
prior alone matches the full pipeline, but as coverage drops its error
grows to 6.8\,mm while the full pipeline remains below 4.5\,mm, so the
layout prior contributes an increasing share of the improvement as the
angular coverage decreases.
Figure~\ref{fig:coverage}B isolates that mechanism: with three full rims
and one hole at 25\% coverage, the radius prior alone yields 6.2\,mm on
the degraded hole against 1.5\,mm on its neighbors, and the
layout projection reduces it to 2.9\,mm at a cost of 0.7\,mm on the
fully covered ones. \rt{The baseline estimates each hole independently}.

Figure~\ref{fig:sweeps} resolves the protocol along its three operating
variables.
Against standoff distance (Fig.~\ref{fig:sweeps}A), the SF baseline
degrades from 7.3\,mm at 1.5\,m to 19.3\,mm at 3.2\,m and
recovers no board at 4.0 and 5.0\,m, where the $0.3^\circ$ pitch spaces
the returns 24 and 31\,mm apart on the board; the full pipeline remains
within
1.8--2.8\,mm through 4.0\,m and reaches 5.0\,mm at 5.0\,m, extending the
usable range of the same board and sensor by more than $1.5\times$, as
the conditioning improvement from $O(\Phi_k^{-4})$ to $O(\Phi_k^{-2})$
predicts.
Against placement (Fig.~\ref{fig:sweeps}B), the lateral positions meet
the LiDAR at $37.5^\circ$, yet the SF full pipeline rises only from 2.4 to
3.7\,mm while the baseline remains between 9.1 and 12.0\,mm, so the
ordering of the variants does not depend on where the board stands in
the image; velo2cam detects 3 of 12 frames at either lateral placement.
Against range noise (Fig.~\ref{fig:sweeps}C), the full pipeline remains
below the baseline at every $\sigma$ on both densities, and on SF frames
it still recovers the board at $\sigma = 50$\,mm, where the baseline has
failed from $\sigma = 35$\,mm.

\begin{table}[t]
\caption{Joint extrinsic in simulation: error against the ground truth, fitting residual, and held-out reprojection}
\label{tab:synth_extrinsic}
\centering
\footnotesize
\renewcommand{\arraystretch}{1.45}
\begin{threeparttable}
\setlength{\tabcolsep}{4pt}
\begin{tabularx}{\linewidth}{|l||>{\centering\arraybackslash}X|>{\centering\arraybackslash}X|>{\centering\arraybackslash}X|>{\centering\arraybackslash}X|>{\centering\arraybackslash}X|}
\hline
\rowcolor{tabhead}
 & & Rot. & Trans. & Joint & LOO \\
\rowcolor{tabhead}
\multirow{-2}{*}{\textbf{Method}} & \multirow{-2}{*}{Det.} & [$^\circ$] & [mm] & [mm] & [px] \\
\hline
\multicolumn{6}{|l|}{\textit{SF}} \\
\hline
velo2cam & 25/60 & \cellcolor{tabbest!25}0.137 & \cellcolor{tabbest!45}3.5 & \cellcolor{tabbest!25}9.5 & \cellcolor{tabbest!25}2.28 \\
\hline
FAST-Calib & 120/180 & \cellcolor{tabbest!45}0.061 & \cellcolor{tabbest!70}2.8 & \cellcolor{tabbest!8}12.8 & \cellcolor{tabbest!8}2.61 \\
Ours w/o RP & 177/180 & \cellcolor{tabbest!70}0.031 & \cellcolor{tabbest!25}4.7 & \cellcolor{tabbest!45}8.6 & \cellcolor{tabbest!45}1.09 \\
Ours w/o LP & 177/180 & \cellcolor{tabbest}\textbf{0.017} & \cellcolor{tabbest}\textbf{0.6} & \cellcolor{tabbest!70}5.1 & \cellcolor{tabbest!70}0.57 \\
\hline
Ours (Full) & 177/180 & \cellcolor{tabbest}\textbf{0.017} & \cellcolor{tabbest}\textbf{0.6} & \cellcolor{tabbest}\textbf{4.6} & \cellcolor{tabbest}\textbf{0.40} \\
\hline
\multicolumn{6}{|l|}{\textit{AC}} \\
\hline
velo2cam & 28/60 & \cellcolor{tabbest!25}0.066 & \cellcolor{tabbest!70}1.9 & \cellcolor{tabbest!8}8.8 & \cellcolor{tabbest!8}1.95 \\
\hline
FAST-Calib & 180/180 & \cellcolor{tabbest!45}0.018 & \cellcolor{tabbest!45}3.0 & \cellcolor{tabbest!25}6.9 & \cellcolor{tabbest!25}1.51 \\
Ours w/o RP & 180/180 & \cellcolor{tabbest!70}0.011 & \cellcolor{tabbest!25}3.3 & \cellcolor{tabbest!45}2.8 & \cellcolor{tabbest!45}0.51 \\
Ours w/o LP & 180/180 & \cellcolor{tabbest}\textbf{0.005} & \cellcolor{tabbest}\textbf{0.3} & \cellcolor{tabbest!70}2.2 & \cellcolor{tabbest!70}0.29 \\
\hline
Ours (Full) & 180/180 & \cellcolor{tabbest}\textbf{0.005} & \cellcolor{tabbest}\textbf{0.3} & \cellcolor{tabbest}\textbf{2.0} & \cellcolor{tabbest}\textbf{0.23} \\
\hline
\end{tabularx}
\begin{tablenotes}[flushleft]
\item \scriptsize\textbf{Note:} The extrinsic is fitted on the detected frames of each variant. Joint is a fitting residual, reported for completeness; LOO holds out one frame. Shading as in Table~\ref{tab:ablation}.
\end{tablenotes}
\end{threeparttable}
\end{table}

\begin{figure*}[!t]
\centering
\setlength{\abovecaptionskip}{1pt}
\includegraphics[width=\textwidth]{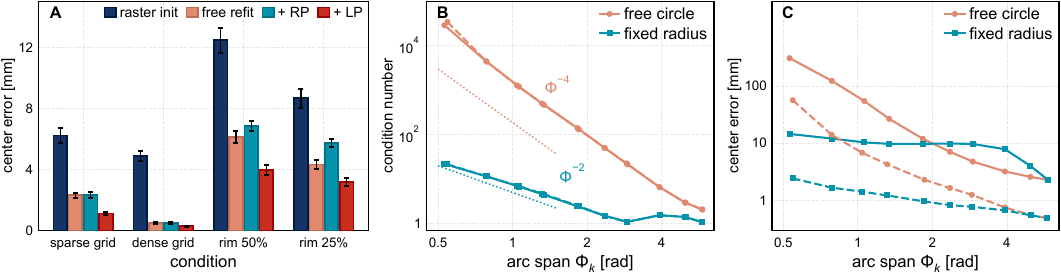}
\caption{Error propagation on simulated boards. (\textbf{A}): hole-center
error after the four stages of Section~\ref{sec:exp_stages}, with 95\%
confidence intervals over 120 holes. (\textbf{B}): condition numbers of
$\tilde{\bm{H}}_k$ and $\bm{H}_k$ as each hole is restricted to $n$
consecutive sectors, SF solid and AC dashed, with the $\Phi_k^{-4}$ and
$\Phi_k^{-2}$ laws of~\eqref{eq:cond_gain} dotted. (\textbf{C}): center
error of the free-circle and fixed-radius estimates over the same sweep.}
\label{fig:stages}
\end{figure*}

\subsubsection{Stage-wise error analysis}
\label{sec:exp_stages}

We trace the hole-center error through the stages of
Algorithm~\ref{alg:p2calib} on centered simulated boards at 3.2\,m and
measure, on the same samples, the condition numbers and error laws that
Section~\ref{sec:theory} predicts.
A free-circle refit on the retained sector samples $\mathcal{E}_k$ is
inserted between the raster initialization and the radius-prior estimate,
so that the effect of the fixed radius is separated from that of the
curated samples.

Figure~\ref{fig:stages}A follows the four stages on SF and AC boards.
Sector sampling provides the largest reduction relative to the raster
initialization, 63\% and 90\%; the fixed radius does not increase the
error at full coverage; and the layout projection removes a further 43
to 52\% of it.
The measured ratio of squared center error after and before the
projection is 0.25 and 0.31, consistent with the $3/8$ predicted
by~\eqref{eq:proj_reduction}, and rises to 0.37 and 0.40 under 25\% and
50\% coverage, where the isotropy assumption weakens.
The conditioning measurements in Fig.~\ref{fig:stages}B follow the two
scaling laws of~\eqref{eq:cond_gain} as each hole is restricted to $n$
consecutive sectors: the free-circle condition number grows over four
decades with a log-log slope of $-4.0$ against the predicted
$O(\Phi_k^{-4})$, while the fixed-radius condition number \rt{remains} below 22
and follows $\Phi_k^{-2}$ once the arc is short.
Figure~\ref{fig:stages}C also identifies the operating limit of the
radius prior: on SF boards with one-sided arcs of 10--12 sectors the free fit
is the more accurate, at 7.1 against 9.9\,mm, because the free
radius absorbs a radial bias that the shared $\delta$ cannot represent.
The layout projection recovers the advantage in each such case
\rt{(Section~\ref{sec:theory_joint})}.


\begin{figure*}[!t]
\centering
\includegraphics[width=0.95\textwidth]{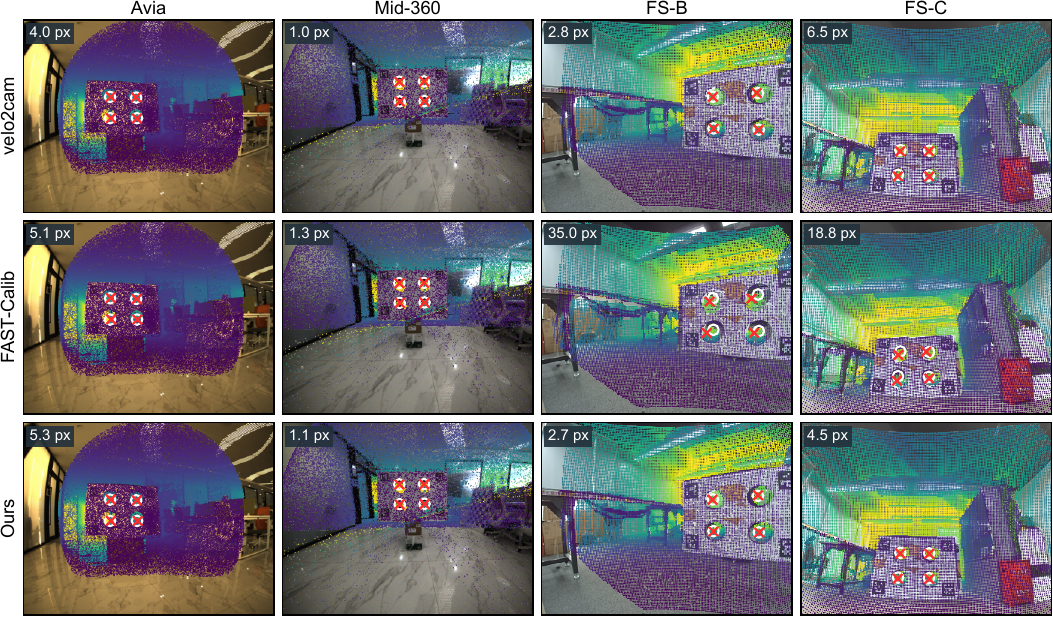}
\caption{Whole-scene projections of the raw scan with the joint extrinsic
of each method. Points are colored by range (viridis, near dark to far
bright); white circles mark the camera-detected hole centers, red crosses
the projected LiDAR centers, and labels the mean center reprojection error
of the shown scene. The disc in the Avia column is the $70^\circ$ conical
field of view of the sensor.}
\label{fig:projection}
\vspace{-0.1cm}
\end{figure*}

\subsection{Real-World Experiments}
\label{sec:exp_real}

\subsubsection{Area-array LiDAR calibration}
\label{sec:exp_fs}

Table~\ref{tab:real} presents the main real-world comparison on the
FS-B and FS-C datasets, which represent the sparse-observation
regime where the two priors are most needed.
Paired errors \rt{are computed on} scenes detected by every variant:
15 in FS-B, 10 in FS-C.

On FS-B, the full pipeline reduces joint residual from 217.7 to
22.3\,mm and LOO error from 67.5 to 2.82\,px, reductions of 90\% and
96\%.
The radius-only variant reaches 24.0\,mm and 3.35\,px, and the
layout-only variant reaches 22.2\,mm and 2.87\,px: both improve
over the baseline but neither matches the full pipeline in both
metrics simultaneously, mirroring the simulated ablation.
On FS-C, joint residual drops from 191.5 to 35.1\,mm and LOO error
from 43.3 to 9.91\,px, reductions of 82\% and 77\%.
The FS-C LOO error is dominated by five scenes at 1\,m standoff,
where every variant including the baseline exceeds 13\,px, with the
centers sitting a consistent $\approx 15$\,mm from the joint
extrinsic in the image plane.
This systematic pattern indicates a range-dependent sensor
inconsistency; excluding these scenes, the full pipeline achieves
4.67\,px on the five scenes at 2--4\,m, consistent with the FS-B
performance.
Figure~\ref{fig:projection} shows representative whole-scene
projections for all four sensors and all three methods.

\begin{table}[t]
\caption{Joint residual [mm] $\downarrow$ and LOO reprojection error [px] $\downarrow$ on the FS datasets}
\label{tab:real}
\centering
\footnotesize
\renewcommand{\arraystretch}{1.45}
\begin{threeparttable}
\setlength{\tabcolsep}{3pt}
\begin{tabularx}{\linewidth}{|l||>{\centering\arraybackslash}X|>{\centering\arraybackslash}X|>{\centering\arraybackslash}X||>{\centering\arraybackslash}X|>{\centering\arraybackslash}X|>{\centering\arraybackslash}X|}
\hline
\rowcolor{tabhead}
 & \multicolumn{3}{c||}{\textbf{FS-B (18 scenes)}} & \multicolumn{3}{c|}{\textbf{FS-C (20 scenes)}} \\
\cline{2-7}
\rowcolor{tabhead}
\multirow{-2}{*}{\textbf{Variant}} & Det. & Joint & LOO & Det. & Joint & LOO \\
\hline
velo2cam\tnote{a} & 14/18 & 23.50 & 3.51 & 11/20 & 40.55 & 10.25 \\
\hline
FAST-Calib & 85/90 & \cellcolor{tabbest!10}217.68 & \cellcolor{tabbest!10}67.50 & 65/100 & \cellcolor{tabbest!10}191.50 & \cellcolor{tabbest!10}43.34 \\
Ours w/o LP & 90/90 & \cellcolor{tabbest!30}24.02 & \cellcolor{tabbest!30}3.35 & 85/100 & \cellcolor{tabbest!30}36.82 & \cellcolor{tabbest!30}10.61 \\
Ours w/o RP & 80/90 & \cellcolor{tabbest}\textbf{22.23} & \cellcolor{tabbest!60}2.87 & 60/100 & \cellcolor{tabbest!60}35.92 & \cellcolor{tabbest}\textbf{9.67} \\
\hline
Ours (Full) & 80/90 & \cellcolor{tabbest!60}22.28 & \cellcolor{tabbest}\textbf{2.82} & 60/100 & \cellcolor{tabbest}\textbf{35.13} & \cellcolor{tabbest!60}9.91 \\
\hline
\end{tabularx}
\begin{tablenotes}[flushleft]
\item \scriptsize\textbf{Note:} Det. counts successful attempts over five seeds; errors \rt{are computed on} the 15 common scenes of FS-B and 10 common scenes of FS-C. Shading as in Table~\ref{tab:ablation}.
\item[a] \scriptsize Its own 14 and 11 detected scenes, hence unshaded.
\end{tablenotes}
\end{threeparttable}
\end{table}

Two observations qualify these results.
First, the shared bias $\delta$ remains at its lower bound of zero on
every real scene of both datasets: the weighted rim radius observed on
the FS boards exceeds the nominal radius by 4--28\,mm, so the radius
prior acts through the fixed radius alone, a sensor-dependent behavior
that Section~\ref{sec:limitations} revisits.
Second, FAST-Calib, RP-only, LP-only, and the full pipeline detect
four centers in 85, 90, 80, and 80 of 90 attempts on FS-B, and in
65, 85, 60, and 60 of 100 attempts on FS-C.
The candidate screening of the layout prior rejects configurations that
the baseline accepts: of the eight FS-C scenes the full pipeline loses,
three provide fewer than four candidates and five fail the 50\,mm
layout check, so detection rate is traded for extraction quality.
The paired results therefore support improved calibration consistency
conditional on successful extraction.

\begin{figure}[t]
\centering
\includegraphics[width=\linewidth]{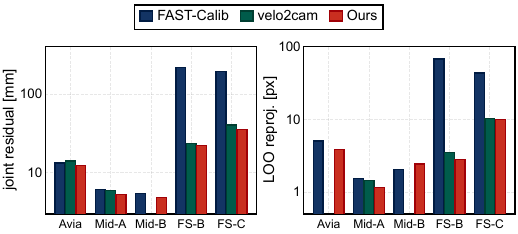}
\caption{Joint residual (\textbf{left}) and LOO error (\textbf{right}) on Avia,
two Mid-360 and FS datasets. FS errors \rt{are computed on} the common
scene sets of Table~\ref{tab:real}, velo2cam on its own detected scenes.}
\label{fig:sensors}
\vspace{-0.1cm}
\end{figure}

\subsubsection{Cross-sensor generality}
\label{sec:exp_sensors}

We verify that the two priors generalize beyond the area-array sensor
to scanning LiDARs with denser angular coverage.
Figure~\ref{fig:sensors} places the two FS datasets of
Table~\ref{tab:real} alongside Avia and the two Mid-360 sessions on a
common scale.

The full pipeline lowers joint residual from 13.2 to 12.4\,mm on
Avia, from 6.1 to 5.3\,mm on Mid-360~A, and from 5.4 to 4.8\,mm on
Mid-360~B.
LOO error decreases from 4.99 to 3.83\,px on Avia and from 1.54 to
1.15\,px on Mid-360~A, but increases slightly from 2.03 to 2.41\,px
on Mid-360~B, indicating that lower joint residual does not guarantee
lower held-out error in every case.
The improvements are smaller than on the FS sensor, as
expected from the conditioning analysis of
Section~\ref{sec:theory_rp}: scanning LiDARs provide denser rim
coverage, yielding larger $\Phi_k$ and a smaller gap between the
$O(\Phi_k^{-4})$ and $O(\Phi_k^{-2})$ curves.
The priors still improve performance, but their marginal contribution
diminishes as the measurement geometry improves.
With the adapted input, velo2cam recovers four centers on 14/18
FS-B, 11/20 FS-C, 2/5 Avia, and 4/4 Mid-360 scenes.
On the detected scenes its joint and LOO errors exceed those of the
full pipeline on both FS datasets \rt{(Table~\ref{tab:real})} while
covering fewer scenes.

\begin{table}[t]
\caption{Runtime [ms] $\downarrow$ of the calibration stages per input cloud}
\label{tab:runtime}
\centering\footnotesize
\renewcommand{\arraystretch}{1.45}
\begin{threeparttable}
\setlength{\tabcolsep}{4.5pt}
\begin{tabular}{|l||l|c||r|r|r||r|}
\hline\rowcolor{tabhead}
\textbf{Method} & \textbf{Dataset} & $n$ & Plane & RP & LP & \textbf{Total} \\
\hline
FAST-Calib & \multirow{2}{*}{FS-B} & \multirow{2}{*}{18} & 7.2 & -- & -- & \cellcolor{tabbest}\textbf{11.9} \\
Ours &  &  & 7.2 & 5.0 & 0.4 & \cellcolor{tabbest!30}16.4 \\
\hline
FAST-Calib & \multirow{2}{*}{FS-C} & \multirow{2}{*}{20} & 9.4 & -- & -- & \cellcolor{tabbest}\textbf{12.3} \\
Ours &  &  & 9.4 & 4.3 & 0.4 & \cellcolor{tabbest!30}18.6 \\
\hline
FAST-Calib & \multirow{2}{*}{Avia} & \multirow{2}{*}{5} & 345.4 & -- & -- & \cellcolor{tabbest}\textbf{356.7} \\
Ours &  &  & 345.1 & -- & 0.4 & \cellcolor{tabbest!30}360.1 \\
\hline
FAST-Calib & \multirow{2}{*}{Mid-360} & \multirow{2}{*}{4} & 69.1 & -- & -- & \cellcolor{tabbest}\textbf{78.2} \\
Ours &  &  & 69.0 & -- & 0.4 & \cellcolor{tabbest!30}81.0 \\
\hline
\end{tabular}
\begin{tablenotes}[flushleft]
\item \scriptsize\textbf{Note:} Median over the $n$ scenes of five warm repeats, file loading excluded; Total also covers the shared raster initialization and bookkeeping. Shading as in Table~\ref{tab:ablation}; a dash marks a stage not executed.
\end{tablenotes}
\end{threeparttable}
\end{table}

\subsubsection{Runtime}
\label{sec:exp_runtime}

Table~\ref{tab:runtime} reports the feature-extraction time measured
on a single thread.
The full pipeline takes 16.4 and 18.6\,ms on FS-B and FS-C, of which
the layout projection accounts for less than 0.5\,ms and the radius
refinement for 4.3 to 5.0\,ms, a 38 and 51\% increase over the
baseline total of 11.9 and 12.3\,ms.
On Avia and Mid-360 plane extraction dominates the total and that
overhead is negligible.

\subsection{Limitations}
\label{sec:limitations}

The current implementation is tailored to the four-hole calibration
board with a rectangular layout.
The radius prior applies to any target with circular features of
known dimension, and the layout prior generalizes to any rigid
arrangement with a known geometric pattern.
The shared bias $\delta$ models inward rim erosion from mixed
pixels; on sensors where the observed radius exceeds the nominal
value, as the FS datasets show, the bias saturates at zero
and the radius prior acts through the fixed radius alone.
Allowing a signed bias or a sensor-specific erosion model may
improve performance on such sensors.
\rt{Whether the same construction helps other target-based estimation
problems whose target geometry is specified but left unused during
fitting remains open.}

\section{Conclusion}
\label{sec:conclusion}

This paper presented P$^2$Calib, a target-based LiDAR--camera
extrinsic calibration system that exploits board geometry priors
to achieve accurate hole-center extraction under sparse point
clouds.
\rt{We showed that fixing the hole radius to its CAD value eliminates the
center--radius degeneracy under sparse angular coverage.
We then enforced the rectangular layout of the four holes as a global
rigidity constraint, which corrects residual errors across holes.}
Both priors are integrated into an interactive calibration tool
that provides a complete extrinsic calibration pipeline.
On two real solid-state LiDAR datasets, P$^2$Calib lowers the joint
residual by 90\% and 82\% and the held-out reprojection error by 96\%
and 77\% over the raster-circle baseline.
\rt{Future work will extend the pattern priors to multi-circle and
non-rectangular board layouts.}

\section*{Conflict of Interest}
The authors declare that they have no known competing financial interests
or personal relationships that could have appeared to influence the work
reported in this paper.

\ifanonymous\else\cameraacknowledgment\fi

\bibliographystyle{IEEEtran}
\bibliography{refs}
\ifanonymous\else
\makeatletter\def\@IEEEBIOphotodepth{0.88in}\makeatother
\begin{IEEEbiography}[{\includegraphics[width=0.64in,height=0.82in,clip,keepaspectratio]{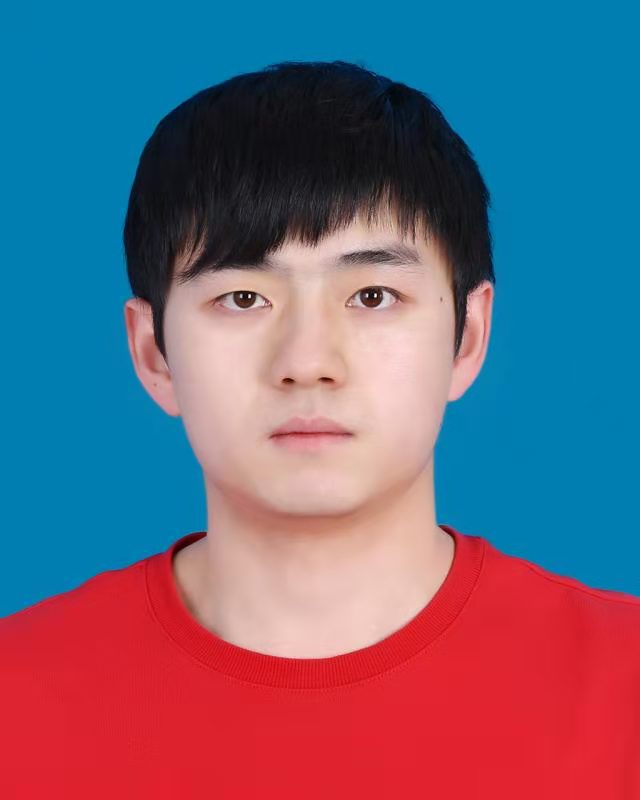}}]{Xiangcheng Hu}
(Student Member, IEEE) received a B.Sc. degree from the North University of China, Taiyuan, China, in 2017 and the M.S. degree from Beihang University, Beijing, China, in 2020. He is currently working toward a Ph.D. degree with the Department of Electronic and Computer Engineering, HKUST.
\end{IEEEbiography}
\fi

\end{document}